**RESEARCH ARTICLE**

# Synthesizing Event-Centric Knowledge Graphs of Daily Activities Using Virtual Space

SHUSAKU EGAMI[1], TAKANORI UGAI[1,2], MIKIKO OONO[1], KOJI KITAMURA[1], AND KEN FUKUDA[1]
[1]National Institute of Advanced Industrial Science and Technology, Tokyo 135-0064, Japan
[2]Fujitsu Ltd., Kawasaki, Kanagawa 211-8588, Japan
Corresponding author: Ken Fukuda (ken.fukuda@aist.go.jp)

This work was supported in part by the New Energy and Industrial Technology Development Organization (NEDO) under Project JPNP20006 and Project JPNP180013, and in part by the Japan Society for the Promotion of Science (JSPS) KAKENHI under Grant JP19H04168 and Grant JP22K18008.

**ABSTRACT** Artificial intelligence (AI) is expected to be embodied in software agents, robots, and cyber-physical systems that can understand the various contextual information of daily life in the home environment to support human behavior and decision making in various situations. Scene graph and knowledge graph (KG) construction technologies have attracted much attention for knowledge-based embodied question answering meeting this expectation. However, collecting and managing real data on daily activities under various experimental conditions in a physical space are quite costly, and developing AI that understands the intentions and contexts is difficult. In the future, data from both virtual spaces, where conditions can be easily modified, and physical spaces, where conditions are difficult to change, are expected to be combined to analyze daily living activities. However, studies on the KG construction of daily activities using virtual space and their application have yet to progress. The potential and challenges must still be clarified to facilitate AI development for human daily life. Thus, this study proposes the *VirtualHome2KG* framework to generate synthetic KGs of daily life activities in virtual space. This framework augments both the synthetic video data of daily activities and the contextual semantic data corresponding to the video contents based on the proposed event-centric schema and virtual space simulation results. Therefore, context-aware data can be analyzed, and various applications that have conventionally been difficult to develop due to the insufficient availability of relevant data and semantic information can be developed. We also demonstrate herein the utility and potential of the proposed VirtualHome2KG framework through several use cases, including the analysis of daily activities by querying, embedding, and clustering, and fall risk detection among older adults based on expert knowledge. As a result, we are able to develop a support tool that detects the fall risk with 1.0 precision, 0.6 recall, and 0.75 F1-score and visualize it with an explanation of its rationale. Using the cases explored in this work, we also clarify and classify the challenges that future research on synthetic KG generation systems should resolve in terms of simulation, schema, and human activity. Finally, we discuss the potential solutions for implementing advanced applications to support our daily life.

## I. INTRODUCTION

Maintaining the safety and quality of life in home environments is becoming increasingly important in an aging society. Our long-term goal is to embody artificial intelligence (AI) in

The associate editor coordinating the review of this manuscript and approving it for publication was Xiaojie Su.

software agents, robots, or cyber-physical systems that can understand the various contextual information of daily life in the home environment to support human behavior and decision making in various situations. Embodied question answering (EQA) [1], which is an AI task that performs visual question answering (VQA) [2] with navigation in a three-dimensional (3D) environment, has recently attracted

IEEE Access    S. Egami et al.: Synthesizing Event-Centric Knowledge Graphs of Daily Activities Using Virtual Spacesignificant attention. In terms of the datasets used for this task, the construction of both scene graphs [3] and knowledge graphs (KG) [4] has attracted particular attention. Knowledge-based EQA [5] has also been proposed. A KG is "*a graph of data intended to accumulate and convey knowledge of the real world, whose nodes represent entities of interest and whose edges represent potentially different relations between these entities*" [4]. So far, various domain-specific KGs, including healthcare, education, ICT, science and engineering, finance, society and politics, and travel, have been used. Abu-Salih surveyed more than 140 papers and provided the following inclusive definition of a domain-specific KG: "*Domain Knowledge Graph is an explicit conceptualisation to a high-level subject-matter domain and its specific subdomains represented in terms of semantically interrelated entities and relations*" [6].

To date, daily activity data are typically collected from physical locations, such as care facilities for older adults [7], smart homes [8], and experimental facilities [9], [10] that imitate living homes. Although some previous studies [11], [12] converted the collected data into KGs, these studies required the use of physical equipment, experimental facilities, and human subjects. Representing a physical environment in virtual space and executing various simulations allow us to freely accumulate the data required for analyzing daily life and constructing EQA/VQA prediction models.

In this study (Figure 1), we focus on cyberspace and propose a method for synthesizing KGs that represent the daily activities in the home environment. Synthesized KGs enable context-aware data analysis and the development of various applications that have previously been difficult to develop due to insufficient data and semantic information.

The primary contributions of this study are presented here. First, we propose the *VirtualHome2KG*[1] framework to generate synthetic KGs based on the simulation results of daily activities using virtual space. We specifically design an event-centric KG schema to represent the daily activities in a home environment. The designed schema can represent object states and properties, affordances, time-series changes, and spatial changes. "Affordance" refers to the action possibilities offered to an animal by the environment in reference to the animal's action capabilities [13], [14]. The proposed VirtualHome2KG framework simulates an arbitrary agent's activities and records the spatiotemporal changes in the virtual space before and after these activities are performed. The simulation results are converted to a KG in the Resource Description Framework (RDF)[2] format based on the designed schema.

Our second contribution is the evaluation of several use cases of the VirtualHome2KG framework. We present several examples of the analysis of daily activity trends, including querying the synthetic KGs and clustering based on a graph embedding method. As an example of a domain-specific application, we focus on fall prevention in the home environment for older adults. We specifically design rules to infer the fall risks for older adults at home and conduct experiments to detect these risks from the generated synthetic KGs. We also developed a support tool for detecting the activities associated with the high fall risk and present explanations about the detected risks. We believe that, in the future, applications detecting accident risks will be helpful tools for residents, home designers, and safety engineering professionals. These applications include cyber-physical systems, AI speakers, and augmented reality systems for injury prevention education [15].

Our third contribution is the organization of the lessons learned from the use cases on the domain-specific applications of synthetic KGs and the discussion of prospects to support human safety using KGs.

Note that this paper is an extension of two previously presented conference papers [16], [17]. The first difference between those papers and this one is that the latter introduced the concept of an event-centric KG to the VirtualHome2KG framework. This schema redesign allowed the addition of risk information to the event nodes that mediated agents, actions, objects, and situations. This redesign also provided a mechanism for presenting the parallel events occurring within an overlapping time span, which was impossible in the previous papers. Furthermore, the redesign made it possible to divert analysis methods similar to existing event-centric KGs [18], [19], [20] (e.g., graph querying and clustering). Aside from adopting an event-centric schema, we refined other components, including actions, places, object attributes, and affordances, and collected more affordance data based on an existing crowdsourcing method [21].

The second difference between our previous conference papers [16], [17] and this one is that the latter presented an experiment and an evaluation conducted to detect the fall risk in older adults in a home environment as a specific use case of the proposed VirtualHome2KG framework. The results demonstrated that the framework can be combined with arbitrary external knowledge and applied to a specific domain. Finally, we described the lessons learned from several use cases and discussed the KG application to support human safety.

The remainder of this paper is organized as follows: Section II describes the related work; Section III explains the synthetic KG construction; Section IV presents several use cases for the constructed KGs; Section V discusses the study results; and Section VI concludes this study with a summary of the potential future work.

## II. RELATED WORK
### A. ACTIVITY RECOGNITION

Recognizing human activities is required to analyze daily life. Activity recognition approaches can generally be classified into two categories [22]. The first approach employs visual sensing devices (e.g., cameras) and recognizes activity patterns in video data using computer vision technologies.

---

[1] https://github.com/aistairc/VirtualHome2KG
[2] https://www.w3.org/RDF/

23858    VOLUME 11, 2023

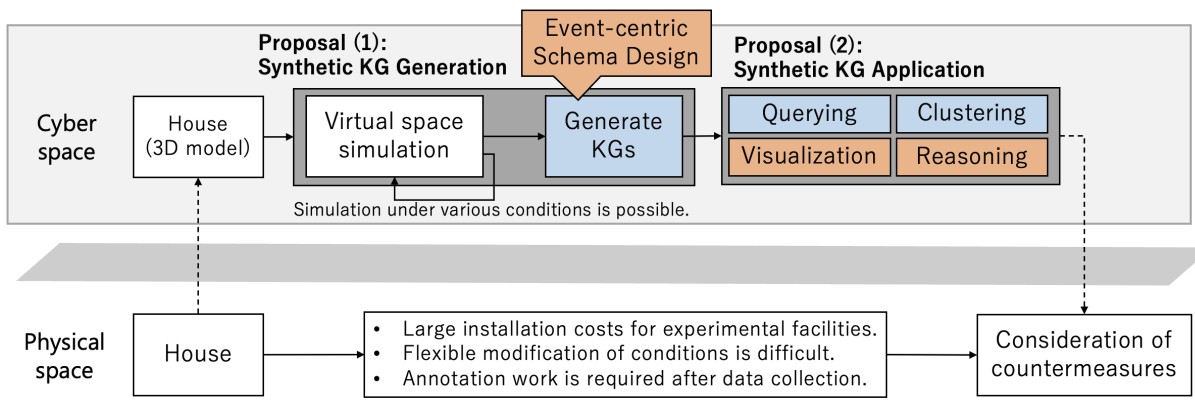

**FIGURE 1.** Overview of the current study. Orange boxes are newly added in this paper, and blue boxes are updated from our previous studies [16], [17].

The second approach monitors human activities using sensor networks and analyzes the data using machine learning techniques. Several data-driven approaches that use machine learning approaches to study the activities of daily living (ADL) include the hidden Markov model, conditional random field [23], [24], and neural network ensemble learning [25]. Several knowledge-driven approaches are also utilized, such as activity recognition based on the ADL ontology [22], and abnormal activity detection using probabilistic logic networks by mapping sensor data to the ontology [26].

Studies related to the construction of scene graph datasets that represent video content (i.e., scenes) as a graph structure have recently increased [3], [27]. A method for improving the activity recognition accuracy by inputting scene graphs with video data into a neural network model has also been proposed [28]. The synthetic KGs generated by the proposed method in this work are similar to the scene graph datasets, in that they represent the video content. However, the scene graphs are typically constructed independently for each video frame. In contrast, our synthetic KGs enable a more context-aware analysis because they have consistent names for nodes and edges and describe the history of situational changes among multiple video frames.

### B. KGs FOR HUMAN ACTIVITY

Several studies constructed KGs based on activity recognition results to improve the ability to analyze daily life. Oh and Jain [11] detected events and then constructed KGs based on the data collected from mobile phones and wearable devices and video data pertaining to the behavior of older adults living in nursing care facilities. Vizcarra et al. [12] constructed KGs based on the object recognition and manual annotation results of video data capturing the behavior of older adults living in nursing homes.

### C. KGs FOR RISK DETECTION

Several studies used KGs to detect risks, such as injuries among children at home and construction site hazards. Oono et al. [15] focused on preventing injuries among children at home and developed an augmented reality lecture system to educate parents on childhood injury prevention. This system recognizes objects from home images obtained from a web camera, refers to a KG representing dangerous situations, and displays information about measures that can be taken to prevent accidents. Oono et al. also constructed KGs of the dangerous relative position information between objects and types of accidents that can occur due to such positions. In contrast, our proposed method constructs event-centric KGs based on human activity simulation data at home and combines the constructed KGs with rules related to fall risk. Our KGs also contain time-series changes in addition to the relative position information. Thus, the proposed method can potentially detect more risks by considering contextual information.

Fang et al. [29] constructed a KG for recognizing hazards on construction sites similar to the fall hazards for older adults in their home environments. They designed an ontology based on engineering documents, accident reports, expert experience, and safety codes, and then they constructed a KG by extracting knowledge from computer vision approaches. As a result, it became possible to identify hazards by querying and reasoning the graph database.

In contrast, the proposed VirtualHome2KG framework can generate KGs containing more spatiotemporally detailed data without being affected by the accuracy of the computer vision method because environmental data (e.g., positions and states of the human body and surrounding objects) are obtained from a 3D virtual space simulator. The proposed framework has high extensibility because it can generate synthetic KGs without limiting the application to a specific domain.

Mao et al. [30] focused on the evolutionary patterns of breaking news events (e.g., geological hazards, traffic accidents, and personal injury) and proposed an event prediction model based on evolutionary event ontology knowledge (EEOK). An EEOK graph was constructed by extracting news events from text data. Our proposed KG contains more spatially detailed data because it is constructed based on the 3D virtual space simulation results.

### D. VIRTUAL SPACE SIMULATION

The aforementioned studies required cameras or sensor devices to focus on physical space. However, various virtual space simulators for embodied AI have since been developed [31], (e.g., VirtualHome [32] enabling the agent simulation of daily household activities and SIGVerse [33], enabling human-robot interactions using virtual reality (VR). To the best of our knowledge, no previous studies have attempted to provide ontology-based semantic labels to the data generated via virtual space simulation to enable an advanced analysis of daily life. Bates et al. [34] proposed a method for recognizing the daily activities obtained from a VR system in real-time and classified the recognized activities based on ontology. Their study primarily focused on activity recognition; thus, knowledge representation and accumulation methods for data exploitation were not considered. Vassiliades et al. [35] constructed HomeOntology based on the activity dataset provided by VirtualHome [32]. This ontology defined household activities (e.g., "relax on the sofa" and "make coffee") comprising multiple primitive actions (e.g., "walk" and "sit") and corresponding target objects. It also defined the activity hierarchy based on the activity knowledge base categories.[3] However, it did not address situations or spatiotemporal information in the home environment. In the current study, we extended HomeOntology to represent various semantic information in the home and daily human activities. The proposed framework generates KGs based on the ontology using virtual space.

Meanwhile, Noueihed et al. [36] developed a virtual outdoor weather event simulator (VOWES) for 3D visualizations. They constructed KGs that describe weather data and 3D simulation components and connected these using semantic sensor network ontology. In contrast, as an event-centric KG, the proposed VirtualHome2KG generates daily household activity simulation data, including human-object interactions. In other words, VOWES and the proposed VirtualHome2KG differ in terms of the target domain, KG structure, and spatial and temporal granularities.

Several studies also reported on the improvement of the accuracy of several real-world tasks using data in virtual space. Miyanishi et al. [37] proposed an approach for improving the accuracy of real-world question answering tasks using the data obtained from a life simulation game. Hwang et al. [38] developed a simulation platform that focused on the daily activities of older adults and used synthetic data to improve the action recognition accuracy. Similar to the proposed VirtualHome2KG framework, existing methods obtain the simulation data from a virtual space. However, they have different objectives and do not focus on enriching semantic information. Nonetheless, they demonstrate that the simulation data obtained from virtual space are effective for real-world tasks. Thus, we believe that the proposed VirtualHome2KG framework can also be applied to such real-world tasks in the future.

## III. CONSTRUCTING SYNTHETIC KGs OF DAILY ACTIVITIES

Figure 2 shows a schematic diagram of our proposed system's data flow and processing modules. This section describes the parts of the source data and synthetic KG generation.

### A. DAILY ACTIVITY SIMULATION USING VIRTUAL SPACE

We used the VirtualHome [32] platform to simulate the daily living activities in a 3D virtual space. Here, the activity data referred to as a *Program* in VirtualHome are represented as a sequence of "steps", comprising an action, an object name, and an object ID. Consider the following example:

```
1   Watch movie
2   Sit down on a couch in front of the TV. Use
        remote to turn on the TV.
3   [WALK] <home_office> (267)
4   [WALK] <couch> (275)
5   [SIT] <couch> (275)
6   [WALK] <remote_control> (1000)
7   [GRAB] <remote_control> (1000)
8   [WALK] <television> (297)
9   [SWITCHON] <television> (297)
10  [TURNTO] <television> (297)
11  [WATCH] <television> (297)
```

The first line identifies the activity name. The second line describes the corresponding activity. In this context, "activity" refers to a coarse-grained event, while "action" refers to a fine-grained event constituting the activity. Each object has a state. For example, a sofa (<couch>) has a state in which nothing is placed before the "sit" action.

The virtual indoor environment was represented by a JSON-formatted graph comprising objects (i.e., nodes in the graph) and their positional relationships (i.e., edges in the graph). Activities can be simulated by loading the object state graph and *program* using the VirtualHome's Unity simulator API. We obtained the crowdsourced activity dataset from VirtualHome's website[4] and used it to perform a simulation in VirtualHome. The home situation was output in JSON format as each action in the activity was performed. For example, the agentcoordinates were updated when the command "[WALK] <home_office> (267)" was executed. The television state changed from OFF to ON when the command "[SWITCHON] <television> (297)" was executed. This way, we recorded the indoor situation when each action in the activity was executed. We also recorded the execution time of each action.

The motions of the most common atomic actions (e.g., "grab" and "walk") were implemented in the VirtualHome Unity simulator. However, the activity dataset in VirtualHome was collected via crowdsourcing and included unimplemented atomic actions (e.g., "wash" and "squeeze"). Note that the Unity simulator cannot render these atomic actions; thus, VirtualHome provides a Graph Evolve Simulator that verifies the executability of the input *program* and updates the object states without rendering an animation. However, this Simulator cannot update the coordinates when an agent or object moves and cannot record each atomic action's execution time. Thus, we extended VirtualHome by combining the Unity and Graph Evolve simulators to record the objects' 3D coordinates and states. We also implemented a function for recording the execution time of each atomic action. The environment graph was acquired after executing each action in the activity. The environment graph list[5] was output after all actions has been executed.

---

[3] http://virtual-home.org/tools/explore.html

[4] http://virtual-home.org/tools/explore.html

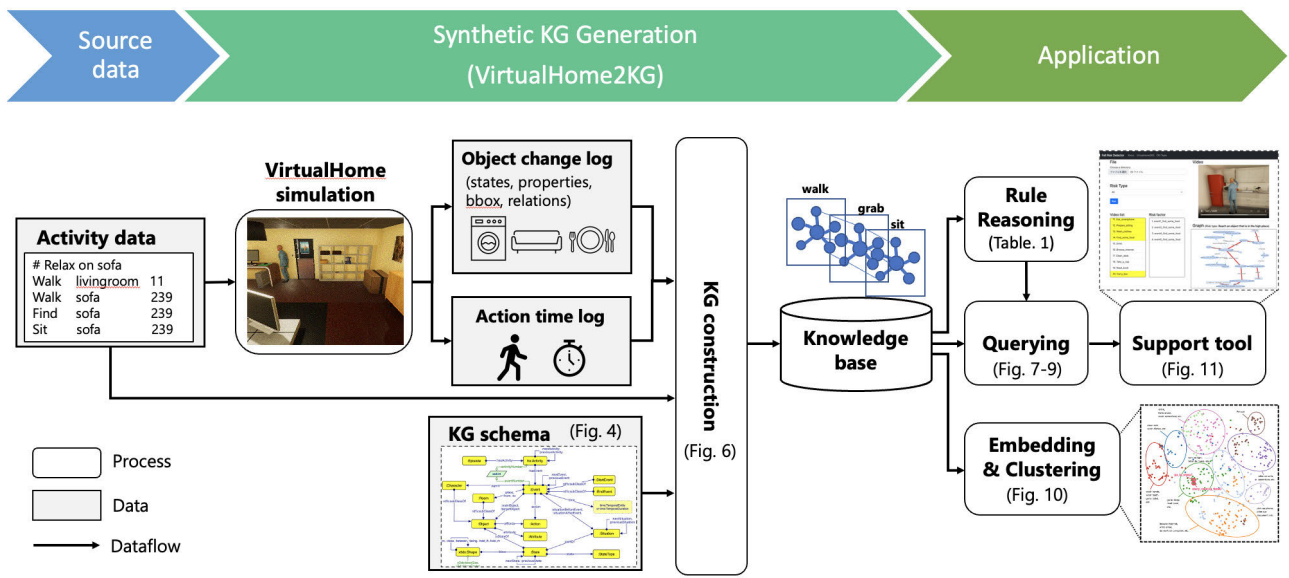

**FIGURE 2.** Schematic diagram of the proposed system.

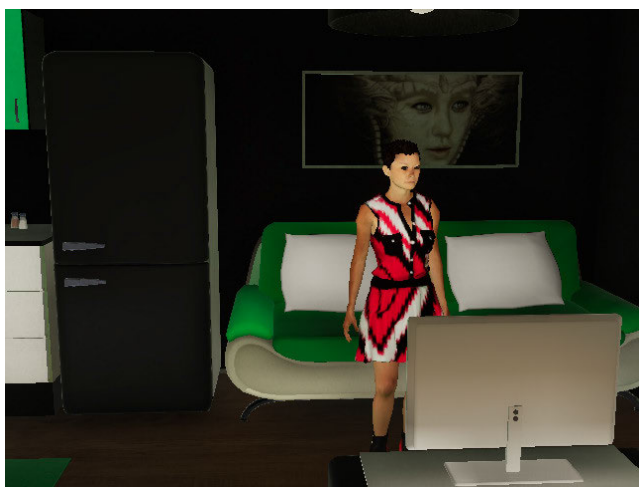

**FIGURE 3.** Example of the execution screen of Unity simulator.

### B. DATA COLLECTION

We used the data collected in a previous study [32] as the source activity dataset and extracted the data from Virtual-Home as the values for the object types and states, properties, attributes, and spatial relationships.

We also collected affordance data by applying the crowdsourcing methodology proposed by Chao et al. [21] to the objects and actions used in this study. The following question was specifically displayed to Amazon Mechanical Turk[6] workers: "Is it possible to X (action) a Y (home object)?" Here, the worker was required to select one of the following options: "Definitely yes"; "Normally yes"; "Maybe"; "Normally no"; "Definitely no"; "I don't know"; or "Description doesn't make sense or is grammatically incorrect." We obtained answers from five different workers for each question and converted the answers into a score ranging from 5.0 ("Definitely yes") to 1.0 ("Definitely no," "I don't know," and "Description doesn't make sense or is grammatically incorrect"). We employed affordances with an average score of 4.0 or greater.

### C. SCHEMA DESIGN

We designed a KG schema representing the human activities and the situational changes in the home environment and generated synthetic KGs based on the designed schema. Figure 4 summarizes the class relation diagram of the designed ontology.

In our previous work, we defined a daily living activity as an "Activity" comprising a sequence of "Actions." In this work, we used the "Action," "Event," "Activity," and "Episode" concepts to represent everyday life in greater detail. The most fine-grained action here is referred to as "Action," an atomic action in VirtualHome. It included "walk," "sit," "grab," and other actions. In VirtualHome's *program*, the combination of atomic action and object is referred to as a "step." Section III-A presents an example for this. We added various information to the step (e.g., agents, actions, places, time, and situations) by representing the ADL as an event-centric KG. In this context, an "Activity" comprises multiple "Events," while an "Episode" comprises multiple "Activities," such as morning routines.

---

[5] Note that the environment graph is not a KG.

[6] https://www.mturk.com/

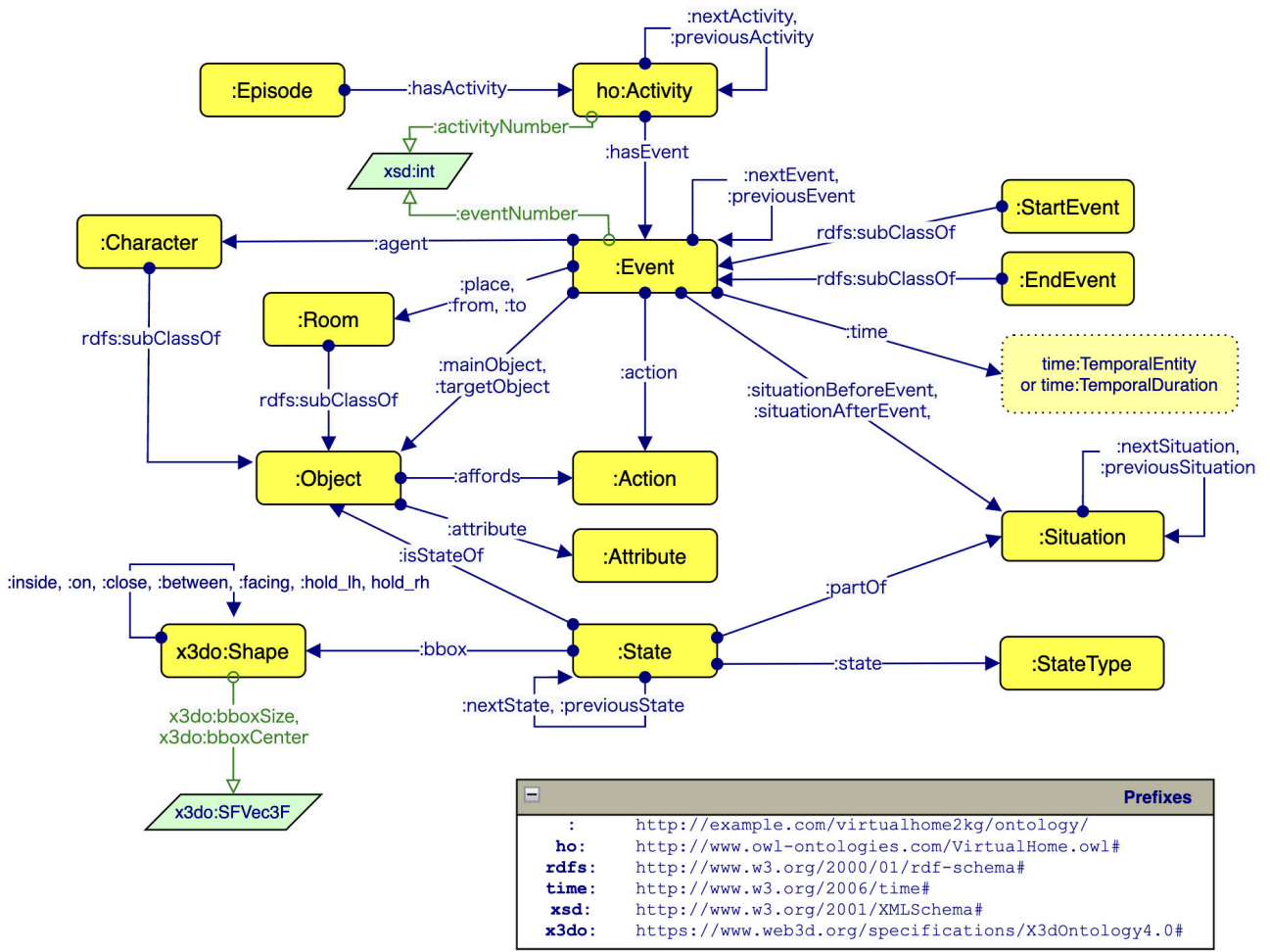

**FIGURE 4.** Graffoo representation [39] of the proposed knowledge graph (KG) schema.

We used the HomeOntology[7] [35] *Activity* class to represent activities. However, HomeOntology does not support the representation of spatiotemporal information; thus, we cannot represent the time required to execute an action and the corresponding 3D coordinates. We also cannot represent the objects' states, affordances, and attributes. The first problem was solved using the Time Ontology[8] and X3D ontology[9] [40]. For the second problem, we defined new classes and properties for these representations.

#### 1) ACTIVITY, EVENT, AND ATOMIC ACTION
The following 12 subclasses were defined in HomeOntology: *BedTimeSleep*, *EatingDrinking*, *FoodPreparation*, *GettingReady*, *HouseArrangement*, *HouseCleaning*, *HygieneStyling*, *Leisure*, *PhysicalActivity*, *SocialInteraction*, *Work*, and *Other*. An additional of 591 subclasses of these 12 classes were also defined. Note that the dataset provided by VirtualHome corresponded to an instance of the lowest-layer classes.

The action types were extracted from the data collected in a previous study [32]. We also expanded the types based on the Primitive Action Ontology [41] to consider the additional actions that may be implemented in the simulation platform in the future.

HomeOntology [35] represents the relationship between *Activity* and *Step* as a list structure using `rdf:List`. However, the list structure based on `rdf:List`, clearly underperforms in commonly used triplestores [42]. Thus, we used the number-based list model to represent the relationship between `ho:Activity` and `Event`. This model is the most efficient method for representing the list in terms of search performance. In addition, we used the Sequence Ontology Pattern (SOP)[10] model to reduce the difficulty of tracing the time-series changes in the object states and activities.

We made the schema more event-centric by attempting to use the Simple Event Model (SEM) [43] and Event Ontology (EO).[11] However, these vocabularies cannot be used due to domain and range restrictions. Thus, we referred only to the ontology design pattern, in which the event classes were directly connected to an agent, a location, and time. Our original classes and properties, which were similar in usage to the SEM and EO, were linked to the corresponding classes and properties using `skos:closeMatch`.[12]

---

[7]https://github.com/valexande/HomeOntology
[8]https://www.w3.org/TR/owl-time/
[9]https://www.web3d.org/x3d/content/semantics/semantics.html

[10]http://ontologydesignpatterns.org/wiki/Submissions:Sequence
[11]http://motools.sourceforge.net/event/event.html

In HomeOntology, action involving multiple target objects (e.g., "PutBack <object1> <object2>") cannot be represented when defining the relationship between an action and an object. Thus, we defined the `mainObject` property to represent the relationship with the first object and the `targetObject` property to represent the relationship with the remaining objects. Note that these properties are subproperties of `ho:object`.

We also defined `place`, `from`, and `to` to represent the specific location of an event. Here, `place` is used if the room, where the agent is located, is the same at both the beginning and end of the event; otherwise, `from` and `to` are used.

The relationships between an activity and an event are described as follows in the Terse RDF Triple Language (Turtle)[13] syntax using our ontology's schema:

```
1  @prefix ex: <http://example.org/virtualhome2kg
       /instance/> .
2  @prefix : <http://example.org/virtualhome2kg/
       ontology/> .
3  @prefix vh2kg-an: <http://example.org/
       virtualhome2kg/ontology/action/> .
4  ex:brush_teeth0_scene1 a ho:brush_teeth ;
5      rdfs:label "Brush teeth" ;
6      :agent ex:character1_scene1 ;
7      :hasEvent ex:event0_brush_teeth0_scene1,
8          ex:event1_brush_teeth0_scene1,
9          ex:event2_brush_teeth0_scene1,
10         ex:event3_brush_teeth0_scene1,
11         ex:event4_brush_teeth0_scene1,
12         ex:event5_brush_teeth0_scene1,
13         ex:event6_brush_teeth0_scene1,
14         ex:event7_brush_teeth0_scene1 ;
15     :virtualHome ex:scene1 ;
16     rdfs:comment "go to the bathroom and brush
            your teeth" .
17 ex:event7_brush_teeth0_scene1 a :EndEvent ;
18     :action vh2kg-an:pour ;
19     :agent ex:character1_scene1 ;
20     :eventNumber "7"^^xsd:int ;
21     :mainObject ex:tooth_paste1000_scene1 ;
22     :previousEvent ex:
            event6_brush_teeth0_scene1 ;
23     :situationAfterEvent ex:
            home_situation8_brush_teeth0_scene1 ;
24     :situationBeforeEvent ex:
            home_situation7_brush_teeth0_scene1 ;
25     :targetObject ex:toothbrush1001_scene1 .
```

Here, lines 1–16 describe triplets with the teeth brushing activity as a subject. It primarily represents the relationships between brushing teeth and events. Lines 17–25 describe the triplets of `event7` as an example primarily representing the target object, previous event, and situations before and after the event.

---

[12] https://www.w3.org/2004/02/skos/
[13] https://www.w3.org/TR/turtle/

### 2) OBJECT's STATE, PROPERTY, ATTRIBUTE, AFFORDANCE, AND SPATIAL RELATIONSHIP

We created the `Situation` class and defined relations to the `Action` class to represent the spatial changes after executing each primitive action in an activity. We also created the `Object` and `State` classes to represent an object and its state, respectively. Furthermore, We defined the next and previous relationships between the object states using the SOP model. With this structure, we can easily track the changes in an object's state. We also defined a `State` at a specific point as part of a `Situation`, which allowed us to simultaneously obtain the states of all objects at a specific point in the virtual space. The `State` class was linked to the state value (`StateType`) and size and the 3D coordinates of an object (`x3do:Shape`). An `StateType` instance specifically represents an object's state (e.g., *ON*, *CLOSED*, and *CLEAN*). We had 33 types of object states, but it was difficult to simulate all possible states in this study. Thus, the object state types were limited based on the Home Object Ontology[14] [41]. By contrast, the KG schema does not limit these values, and new state types can be added.

VirtualHome also defines the *Object Properties* (e.g., *GRABBABLE*, *HAS_SWITCH*, and *CLOTHES*). In *Object Properties*, both the attributes and the affordances of objects are mixed together. Accordingly, we defined the `Attribute` and `affords` and classified the values of the object properties in VirtualHome. We assumed that affordance is unaffected by an object's state and represents the possible actions given by the object's existence. Therefore, we defined the affordance structure as the relation, in which an `Object` affords an `Action`.

The `x3do:Shape` class from the X3D ontology was used. This class has qualitative spatial relations with other objects.

### D. KG CONSTRUCTION

The VirtualHome simulation results were converted to KGs based on the schema described in Section III-C. Figure 5 shows part of a KG constructed based on the described ontology. This part of the KG results from executing "Listen to music," which is one of the activities provided by the VirtualHome. Note that there are over 400 objects and their corresponding states. Here, an instance of `State` was created only when an object's state, coordinate, or affordance changed before and after an action execution. If the object does not change, even after executing the next primitive action, the instance of `State` is connected to the instances of the current `Situation` and the next `Situation` through the `partOf` property. Thus, many state instances of a frequently changing object (e.g., `Character`) are generated. Conversely, an instance of the state of an object, which is not the target of the activity becomes a part of all the situation instances because the object state is never changed.

Figure 6 shows a flowchart diagram of the KG construction process. We developed two script types to generate the KGs.

---

[14] https://github.com/aistairc/HomeObjectOntology

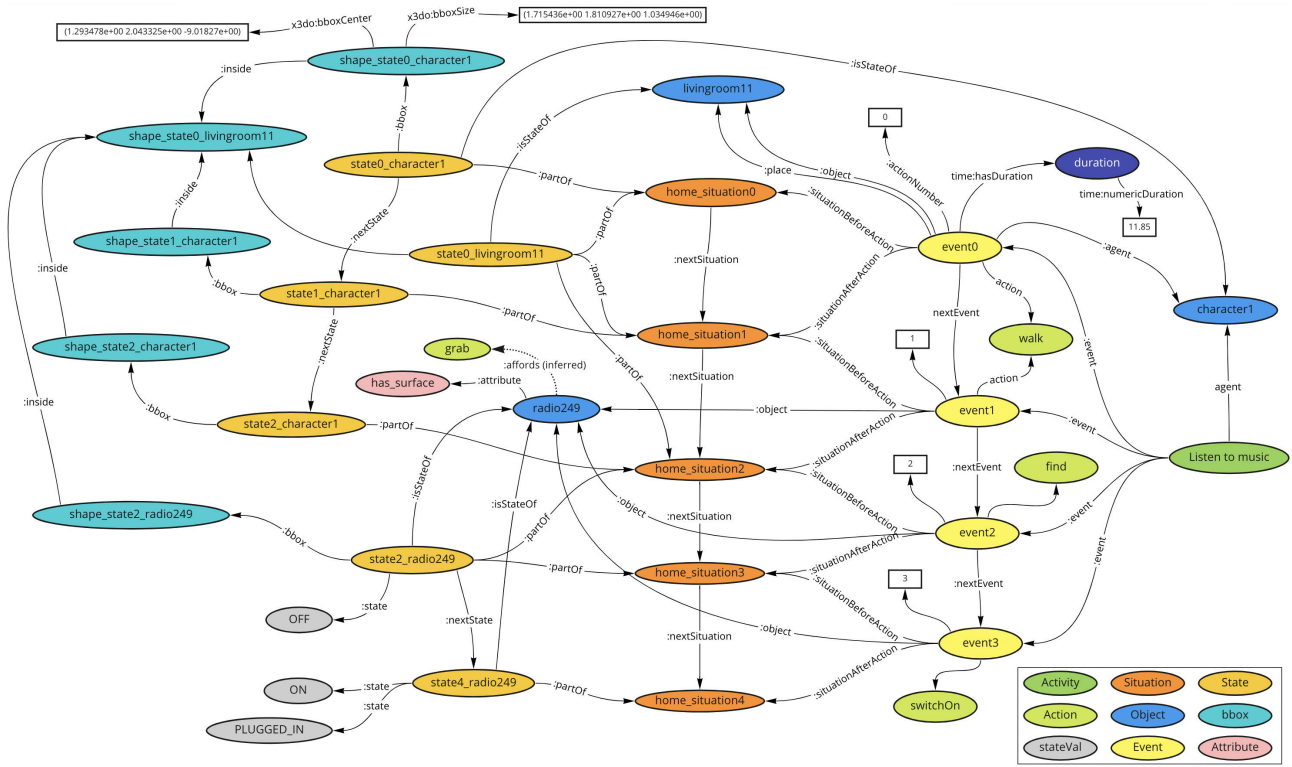

**FIGURE 5.** Part of a constructed knowledge graph.

The first type generated KGs containing 3D coordinates and the duration of action (i.e., execution time) by rendering the *program* using the Unity simulator. Consequently, it took a long time to generate the KGs. The motion data of the atomic actions implemented by the VirtualHome were limited; hence, the simulation's executability was dependent on the VirtualHome Unity simulator. In contrast, the second script type generated KGs without recording the 3D coordinates of objects[15] and the action duration. This script can quickly generate a large amount of data because it performed the simulations without rendering the *program* using the Unity simulator.

We simulated and rendered 43 executable activities in the VirtualHome activity dataset as samples. We then constructed a KG using the designed ontology. The constructed KG and schema data were stored in Ontotext GraphDB,[16] which is a triplestore. The number of entities was 314,662. The number of properties was 54. The number of triples (relationships) was 1,126,740. We also simulated 451 executable activities without rendering, and we constructed a KG based on the designed schema. The number of entities was 1,290,252. The number of properties was 52. The number of triples was 8,477,485. Lastly, the activity included an average of 10.13 primitive actions.

---

[15]Note that the state changes and the qualitative positional relationships of the objects were not removed.

[16]https://www.ontotext.com/products/graphdb/

## IV. USE CASES OF SYNTHETIC KGs

This section describes several use cases to that demonstrate the usefulness of the synthetic KGs generated using the proposed VirtualHome2KG framework. These use cases included the analyses of daily living activities and fall risk detection in a home environment.

### A. ANALYSIS OF DAILY LIVING ACTIVITIES

#### 1) QUERYING USING SPARQL

Various daily activities can be analyzed by querying the generated synthetic KGs using SPARQL. For example, a SPARQL query to search for "frequently grabbed objects[17]" can yield the results shown in Figure 7. This query can help identify the source of indirect contact in an infectious disease context. Figure 8 shows the results of the query for "objects with frequent state changes.[18]" These results allow users to predict the objects' wear and tear. The conditions of these two queries can be combined. Figure 9 shows the results of the query for the "leisure activities that consume a lot of time." The generated KGs contained the execution time for each event in the activities; hence, it was possible to aggregate using the temporal information. Combining this query with the target object and location conditions also allows us to more spatiotemporally analyze daily activities.

---

[17]https://github.com/aistairc/VirtualHome2KG/tree/main/sparql#example-3

[18]https://github.com/aistairc/VirtualHome2KG/tree/main/sparql#example-4

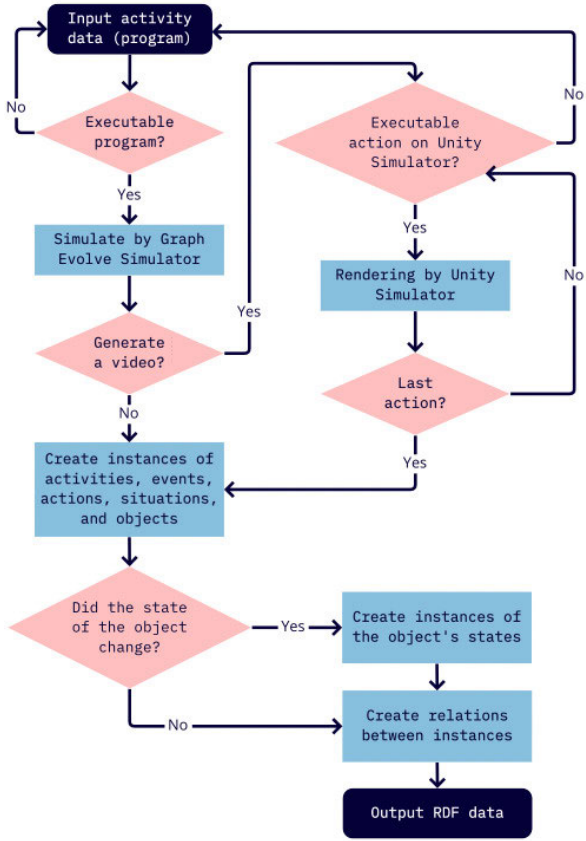

FIGURE 6. Flowchart diagram of KG construction process.

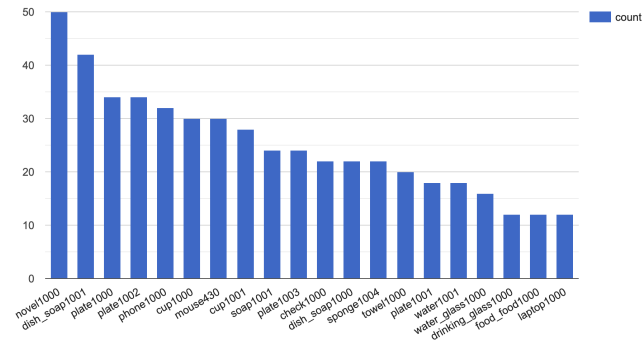

FIGURE 7. Top 20 most frequently grabbed objects.

### 2) VISUAL ANALYSIS BASED ON GRAPH EMBEDDING

The potential semantics can be interpreted by embedding each entity in the KG in a high-dimensional vector space and obtaining a vector representation. We will describe herein how to obtain a vector representation of the Activity instances and visualize them to analyze the features of daily living activities. In this case, we employed RDF2Vec [44] as the graph embedding method because our KG is in an RDF format. RDF2Vec generates a sequence set using a random walk and learns the vector representation using a continuous bag-of-words (CBOW) or a skip-gram model.

First, for a given graph $G = (V, E)$, for each vertex $v \in V$, all graph walks $P_v$ of depth $d$ were rooted in the vertex $v_r$,

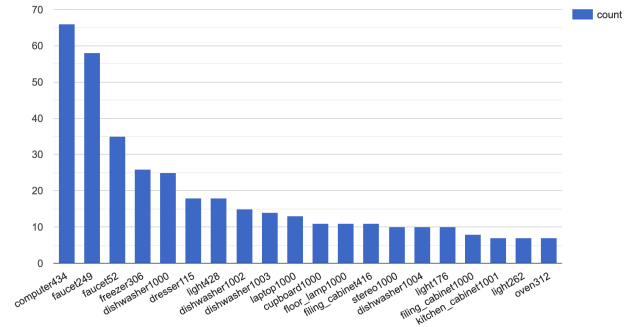

FIGURE 8. Top 20 objects with frequent state changes.

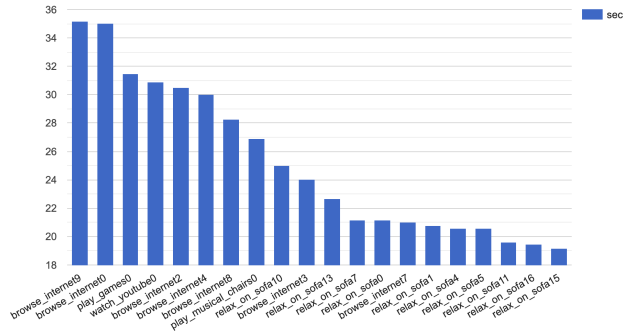

FIGURE 9. Most time-consuming leisure activities.

e.g., $v_r \rightarrow e_{1,1} \rightarrow v_1$, where $i \in E(v_r)$. Then, vertex and edge labels were then attached using the graph labeling method in the Weifeiler–Lehman (WL) graph kernels for RDF [45]. The relabeling process continued until $h$ iterations were reached. Walk paths were then extracted based on the $P_v$ on each relabeling graph. Thus, the final set of sequences was the union of the sequences of all vertices in each iteration $\bigcup_{i=1}^{h} \bigcup_{v \in V} P_v$. To generate the vector representation, the final set of sequences was input to word2vec [46]. We used the skip-gram model because it tends to achieve a high accuracy in RDF2Vec. It is the inverse model of the CBOW. When a word $w_t$ is given, the skip-gram model maximizes the log probability of words $w_{t+j}(-c \leq j \leq c)$ appearing in the window size $c$.

$$\frac{1}{T} \sum_{t=1}^{T} \sum_{j} log(p(w_{t+j}|w_t)) \quad (1)$$

where the probability $p(w_{t+j}|w_t)$ is calculated as follows using the SoftMax function.

$$p(w_o|w_i) = \frac{exp(v'^T_{w_o} v_{w_i})}{\sum_{w=1}^{V} exp(v^T_w v_{w_i})} \quad (2)$$

After embedding the KG into a 100-dimensional vector space using RDF2Vec, we applied the UMAP [47] dimension reduction method to visualize the results as a two-dimensional plot (Figure 10)d. We also classified the activity instances into 10 clusters using the k-means clustering algorithm. We set the cluster size based on the hypothesis that 10 clusters will be generated because this experiment limited the experimental data to 10 activity types, excluding the *PhysicalActivity* and *SocialInteraction* types, which cannot be executed on the simulator. Entities distant in a 100-dimensional space may be placed close to each other in a 2D space through the dimension reduction method. We performed the k-means clustering on a 100-dimensional space to accurately represent this phenomenon. In Figure 10, the meaningful sets were circled for visibility (i.e., the circles do not represent the number of clusters.) As a result, similar activities were plotted close to each other. However, in some cases, activities with the same name were plotted in different positions due to the various combinations of actions and objects comprising the corresponding activity. Conversely, in other cases, activity instances with different names were plotted close to each other. For example, the "go_to_sleep1" and "story_reading_time0" activities were plotted close to each other, as depicted by the magenta circles in Figure 10. The "go_to_sleep1" activity instance represents walking to a bedroom, finding a bed, and then lying down on the bed. The "story_reading_time0" activity instance represents walking to a bedroom, finding a bed, and then sitting on the bed before reading a book. The context of a longer period of time influenced the embedding results (i.e., similarities in their actions, target objects, and places). In addition, the clustering results implied that the class hierarchy of daily household activities may be redifined.

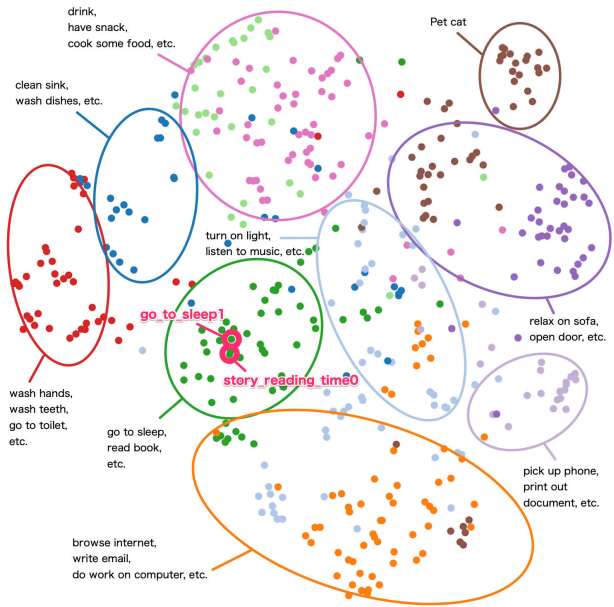

**FIGURE 10.** Visualization of UMAP results of the vectors of the learned *Activity* instances obtained using RDF2Vec (parameter settings: depth = 8, number of walks per entity = 100, WL iterations = 6, vector size = 100, window size = 9, model = skip-gram, skip predicates = {agent, hasActivity, virtualHome (identifying a home), partOf, previousEvent}).

### B. DETECTING FALL RISK EVENTS AT HOME

This section describes an experiment conducted to detect the fall risk among older adults in the home environment as an application of the proposed VirtualHome2KG framework.

#### 1) DEFINING THE FALL RISK CLASSES AND RULES

First, we defined two risk classes (i.e., `RiskActivity` and `RiskEvent`) and the subclasses of `RiskEvent` to infer the high risk behaviors for older adults from the synthetic KGs generated by the proposed VirtualHome2KG framework. We also defined the riskFactor property to represent the risk factors.

Next, one safety engineering expert manually extracted the risk behaviors and situations leading to falls at home from the accident data provided by the Tokyo Fire Department and the video data from the Elderly Behavior Library.[19] The accident data included reports of daily accidents that required patients to be transported by ambulance in Tokyo. These data comprised information about the situation of the accident, related injuries, location, what the patient was doing, and their age and gender. These reports were typically written as a short natural language text. We then classified the extracted risks into the following three categories:

(1) Dangerous action: this action is dangerous, regardless of the target object

- -- Go up or down the steps
- -- Straddle an object
- -- Walk backwards
- -- Stand on one leg
- -- Do some work using one's foot
- -- Stand up without support

(2) Dangerous interaction: the combination of an action and the object's characteristics is dangerous

- -- Reach an object that is in a high place
- -- Take an object out of low shelves
- -- Carry a heavy object
- -- Lean on an unstable object
- -- Pick up an object on the floor while sitting on a chair

(3) Dangerous spatial relationship

- -- An object is placed on an aisle.
- -- There is a gap between the bed and the wall.
- -- A cushion is laid on a chair.
- -- A bed has no side rails.
- -- A chair has no armrest.

In this experiment, duplicate expressions and phenomena were excluded from the classification process. For example, the risk list extracted by the expert included "standing in a high place." However, before standing in a high place, a person should have already climbed something. The risk list already included "go up or down the steps"; thus, we excluded "standing in a high place" from the list because it is semantically redundant. Although the extracted risk list also included "losing balance," it is a phenomenon caused by various factors and not an intentional activity. We aimed herein to detect the risk before the phenomenon occurs, and not the resulting phenomenon. Therefore, we excluded "losing balance" from the risk list.

---

[19]http://www.behavior-library-meti.com/behaviorLib/

**TABLE 1.** Inference rules for the fall risk among older adults at home.

| ID | Fall risk | Alternative expression | SPARQL query corresponding to the inference rule |
|---|---|---|---|
| R1 | Reach an object that is in a high place | Perform actions other than "walk," "turn to," and "look at" on objects taller than one's height | CONSTRUCT {<br>?a hra:riskFactor ?e . ?e a hra:DoSomethingToHighPositionObject .<br>} WHERE {<br>?a :hasEvent ?e . ?e :agent ?person ; :situationBeforeEvent ?situation ;<br>ho:object ?o ; :action ?action . ?o :height/rdf:value ?oh .<br>?person :height/rdf:value ?ph . ?state1 :isStateOf ?person ;<br>:partOf ?situation ; :bbox ?shape1 . ?state2 :isStateOf ?o ; :partOf ?situation ;<br>:bbox ?shape2 . ?shape1 x3do:bboxCenter ?center1 .<br>?center1 rdf:rest/rdf:first ?center_y1 . ?shape2 x3do:bboxCenter ?center2 .<br>?center2 rdf:rest/rdf:first ?center_y2 .<br>FILTER ((?center_y2 + (?oh * 0.5)) >(?center_y1 + (?ph * 0.5)))<br>FILTER (?action != ac:walk && ?action != ac:watch<br>&& ?action != ac:turnTo && ?action != ac:lookAt)<br>MINUS {?o rdf:type/rdfs:subClassOf* :Room }<br>} |
| R2 | Take an object out of low shelves | Grab an object placed lower than the center of one's body | CONSTRUCT {<br>?a hra:riskFactor ?e . ?e a hra:GrabLowPositionObject .<br>} WHERE {<br>?a :hasEvent ?e . ?e :agent ?person ; :situationBeforeEvent ?situation ;<br>ho:object ?o ; :action ac:grab . ?o :height/rdf:value ?oh .<br>?person :height/rdf:value ?ph . ?state1 :isStateOf ?person ;<br>:partOf ?situation ; :bbox ?shape1 . ?state2 :isStateOf ?o ;<br>:partOf ?situation ; :bbox ?shape2 . ?shape1 x3do:bboxCenter ?center1 .<br>?center1 rdf:rest/rdf:first ?center_y1 . ?shape2 x3do:bboxCenter ?center2 .<br>?center2 rdf:rest/rdf:first ?center_y2 .<br>FILTER ((?center_y2 + (?oh * 0.5)) <?center_y1 )<br>MINUS {?o rdf:type/rdfs:subClassOf* :Room }<br>} |

**TABLE 2.** Confusion matrix.

|  |  | Ground truth | |
|---|---|---|---|
|  |  | Risk event | No risk event |
| Detection | Risk event | TP | FP |
|  | No risk event | FN | TN |

We focused on the fall risks among older adults in their home environment caused by an interaction between a behavior and the environment. We detected the two following risks from the synthetic KGs: "reach an object that is in a high place" and "take an object out of low shelves." We modified the fall risks to more specific alternative expressions to map the words in the risk expressions to the entities in the synthetic KGs. We then defined inference rules corresponding to the alternative expressions (Table 1).

2) EVALUATION

We generated videos and KG experimental data using the proposed VirtualHome2KG framework. In this evaluation, an expert verified the videos and annotated the risk scenes using the ELAN[20] annotation tool for the video data. The ground truth included the event nodes corresponding to the scenes labeled to contain a fall risk. The confusion matrix classified events into true positive (TP), false positive (FP), true negative (TN), and false negative (FN) (Table 2).

The precision, recall, and F1-score of the proposed method were calculated as follows:

$$Precision = \frac{TP}{TP + FP} \quad (3)$$

$$Recall = \frac{TP}{TP + FN} \quad (4)$$

$$F1 = \frac{2 \times Recall \times Precision}{Precision + Recall} \quad (5)$$

We initially considered random sampling from the activity dataset of a previous study [32] and generating videos and KGs from the extracted samples. However, with this technique, many activities could not successfully generate videos due to the VirtualHome data limitations. For example, 36% of the activities in the crowdsourced dataset was not executable [32]. Although VirtualHome prepared 3D assets for the simulator, the availability of motion data, 3D models of manipulatable objects, and object gimmicks are not yet sufficient for generating a wide variety of daily life activities for random sampling. Some generated videos also include unnatural behaviors, such as skipping required actions, floating in the air, and exceeding the rotation range of joints due to an insufficiently defined rotation limit of body parts. We eliminated these activity data by preparing the 20 activity scenarios listed in Table 3 to generate videos and KGs that did not contain unnatural scenes. These scenarios were designed as naturally occurring activities in daily life and carefully considered to cover as many activity types as possible. We generated videos and synthetic KGs based on these scenarios and used them in our experiment.

[20]https://archive.mpi.nl/tla/elan

**TABLE 3.** Ground truth dataset for detecting the fall risks.

| Activity (type) | Script | # of event | # of risk event | Risk type |
|---|---|---|---|---|
| Admire paintings (Leisure) | [WALK] <wallpictureframe> (419), [TURNTO] <wallpictureframe> (419), [LOOKAT] <wallpictureframe> (419) | 3 | 0 | - |
| Browse internet (Work or Leisure) | [WALK] <livingroom> (336), [WALK] <computer> (434), [SWITCHON] <computer> (434), [WALK] <chair> (373), [SIT] <chair> (373) | 5 | 0 | - |
| Brush teeth (HygieneStyling) | [WALK] <bathroom> (11), [WALK] <sink> (248), [GRAB] <toothbrush> (66), [SWITCHON] <faucet> (51) | 4 | 0 | - |
| Carry box (HouseArrangement) | [WALK] <box> (194), [GRAB] <box> (194), [WALK] <floor> (337), [TURNTO] <wallpictureframe> (419), [PUTBACK] <box> (194) <floor> (337) | 5 | 1 | R2 |
| Clean desk (HouseCleaning) | [WALK] <desk> (110), [GRAB] <mug> (196), [WALK] <sink> (247), [PUTBACK] <mug> (196) <sink> (247) | 4 | 0 | - |
| Drink water (EatingDrinking) | [WALK] <kitchen> (207), [WALK] <glass> (271), [GRAB] <glass> (271), [WALK] <faucet> (249), [SWITCHON] <faucet> (249), [WALK] <sink> (247), [PUTBACK] <glass> (271) <sink> (247), [SWITCHOFF] <faucet> (249), [GRAB] <glass> (271), [DRINK] <glass> (271) | 10 | 0 | - |
| Find some foods (FoodPreparation) | [WALK] <fridge> (306), [OPEN] <fridge> (306), [CLOSE] <fridge> (306), [WALK] <kitchencabinet> (237), [OPEN] <kitchencabinet> (237), [CLOSE] <kitchencabinet> (237) | 6 | 2 | R1 |
| Go to sleep (BedTimeSleep) | [WALK] <bedroom> (74), [WALK] <light> (175), [SWITCHOFF] <light> (175), [WALK] <bed> (111), [SIT] <bed> (111) | 5 | 0 | - |
| Go to toilet (HygieneStyling) | [WALK] <bathroom> (11), [WALK] <toilet> (46), [SIT] <toilet> (46), [STANDUP], [TURNTO] <toilet> (46), [TOUCH] <toilet> (46) | 6 | 0 | - |
| Prepare breakfast (FoodPreparation) | [WALK] <kitchen> (207), [WALK] <breadslice> (310), [GRAB] <breadslice> (310), [WALK] <plate> (274), [PUTBACK] <breadslice> (310) <plate> (274) | 5 | 0 | - |
| Prepare sitting (HouseArrangement) | [WALK] <bookshelf> (250), [GRAB] <clothespile> (287), [WALK] <chair> (373), [PUTBACK] <clothespile> (287) <chair> (373) | 4 | 1 | R2 |
| Read book (Leisure) | [WALK] <bedroom> (74), [WALK] <bookshelf> (107), [GRAB] <book> (192), [WALK] <sofa> (369), [SIT] <sofa> (369) | 5 | 0 | - |
| Relax on sofa (Leisure) | [WALK] <livingroom> (336), [WALK] <sofa> (369), [WALK] <pillow> (422), [GRAB] <pillow> (422), [WALK] <sofa> (369), [SIT] <sofa> (369) | 6 | 0 | - |
| Take a nap (BedTimeSleep) | [WALK] <bedroom> (74), [WALK] <tablelamp> (104), [SWITCHOFF] <tablelamp> (104), [WALK] <bedroom> (74), [WALK] <bed> (111), [SIT] <bed> (111) | 6 | 0 | - |
| Take off clock (HouseArrangement) | [WALK] <clock> (266), [GRAB] <clock> (266) | 2 | 1 | R1 |
| Use smartphone (Leisure) | [WALK] <bedroom> (74), [WALK] <cellphone> (190), [GRAB] <cellphone> (190), [WALK] <livingroom> (336), [WALK] <sofa> (369), [SIT] <sofa> (369) | 6 | 0 | - |
| Wash clothes (HouseCleaning) | [WALK] <bookshelf> (250), [GRAB] <clothespile> (287), [WALK] <bathroom> (11), [WALK] <washing_machine> (73), [PUTBACK] <clothespile> (287) <washing_machine> (73) | 5 | 1 | R2 |
| Wash hands (HygieneStyling) | [WALK] <bathroom> (11), [WALK] <sink> (248), [SWITCHON] <faucet> (51), [SWITCHOFF] <faucet> (51), [WALK] <towel> (56), [GRAB] <towel> (56) | 6 | 0 | - |
| Wash pillow (HouseCleaning) | [WALK] <pillow> (423), [GRAB] <pillow> (423), [WALK] <washing_machine> (73), [TURNTO] <washing_machine> (73) | 4 | 0 | - |
| Watch tv (Leisure) | [WALK] <livingroom> (336), [WALK] <tv> (427), [SWITCHON] <tv> (427), [WALK] <sofa> (369), [TURNTO] <tv> (427), [SIT] <sofa> (369) | 6 | 0 | - |

**TABLE 4.** Experimental results.

| | | Ground truth | |
|---|---|---|---|
| | | Risk event | No risk event |
| Detection | Risk event | 0.6 (6/10) | 0.4 (4/10) |
| | No risk event | 0 (0/93) | 1.0 (93/93) |
| Accuracy | Precision | 0.6 | |
| | Recall | 1.0 | |
| | F1 | 0.75 | |

Table 4 shows the experimental results. All fall risks targeted in this experiment were correctly detected, and a 0.75 F1-score was achieved. The proposed method is effective because recall is more important than precision in terms of preventing life-threatening injuries among older adults.

Table 4 also shows room for improvement in precision. In other words, some non-risky events were detected as representing fall risks. For example, the event representing yjr opening of a refrigerator door was detected as a "Reach an object that is in a high place" risk. However, this was a false positive because the position, at which the agent reached to open the refrigerator door, was not taller than the agent's height. We believe that these false positives can be solved if the part-whole relations of objects are included (e.g., the relations between the door handle and the refrigerator). Alternatively, risk may be detected by obtaining information describing the arm's position from the skeleton information.

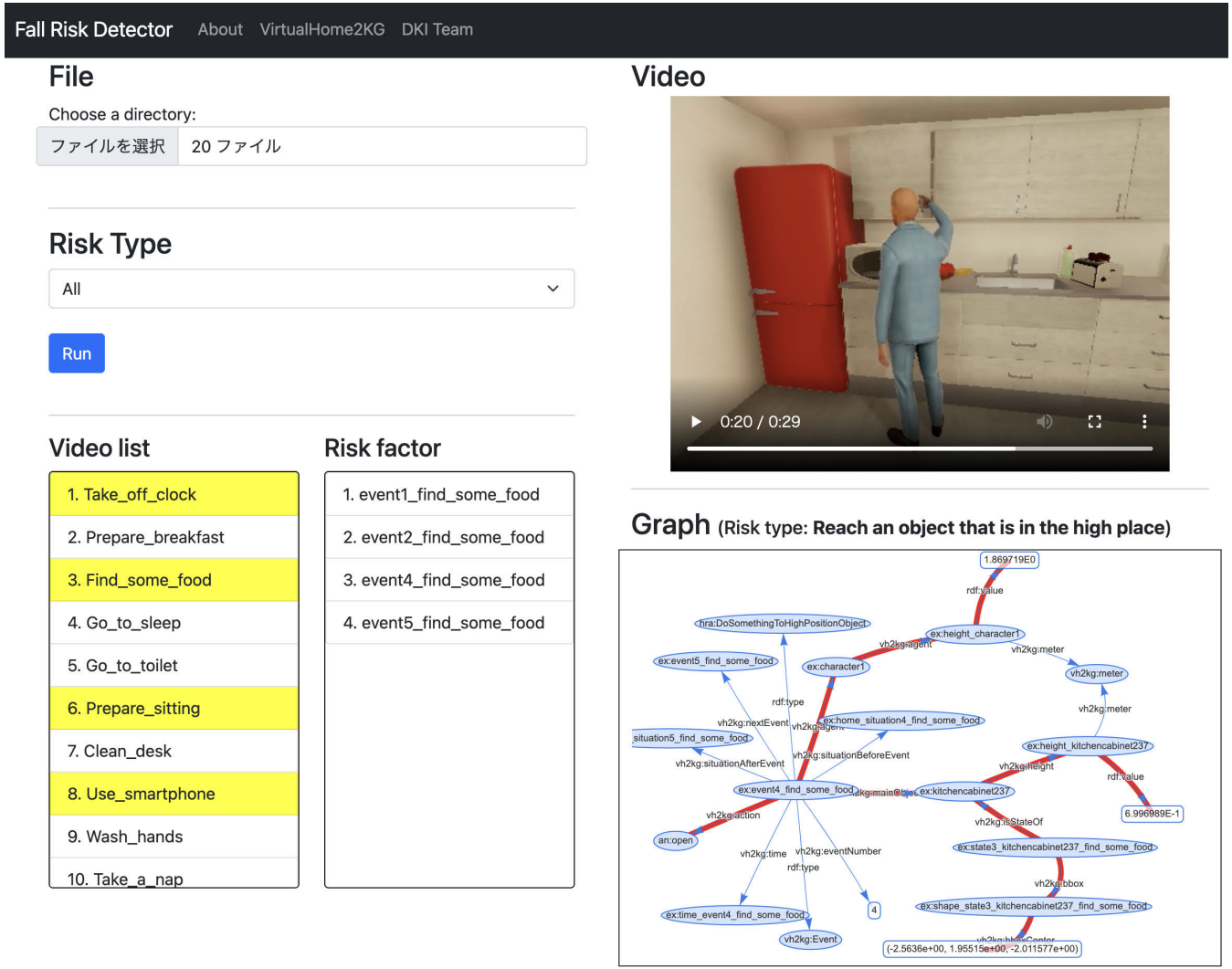

**FIGURE 11.** Support tool to detect and explain fall risks.

#### 3) SUPPORT TOOL
We believe that the simulation and demonstration of risky behaviors for educational purposes, which can help develop countermeasures to reduce accidents in living environments, are promising future applications of explainable AI in the safety domain. To explore the present feasibility, we developed a support tool for fall risk detection (Figure 11). The application loads videos generated by the proposed VirtualHome2KG and detects the fall risks from the synthetic KGs corresponding to the loaded videos. Among the videos displayed in the ''Video list,'' in Figure 11, those judged to contain a fall risk are highlighted in yellow. When users select the risk video, the events judged as a fall risk in the activity are displayed as a ''Risk factor.'' The corresponding video scene is played in a loop the moment an event name is selected. In addition, the KG centered around the event node is drawn; the risk description is displayed to the user; and the graph path considering the fall risk basis is highlighted in red. Accordingly, we can now develop an application with a high explanatory ability that demonstrates the fall risk basis with videos, text content, and graph paths using the data generated by the proposed VirtualHome2KG framework. We believe that this application can help guide the design of serviced housing for older adults.

### V. DISCUSSION
VirtualHome2KG enables the generation of a large number of synthetic KGs by simulating daily life in a virtual space. We believe that the proposed approach contributes to increasing the knowledge captured and embodied in AI systems to support daily life.

Table 5 compares the proposed VirtualHome2KG framework with the existing methods that use VirtualHome. VirtualHome2KG has a more efficient querying ability than existing methods. In particular, it can capture the context of daily behavior and the environment, depicting a potential application in both VQA and EQA systems. The proposed VirtualHome2KG framework can also represent more knowledge than HomeOntology and has particular strengths in terms of the knowledge representation of the spatial, contextual, and property relationships focusing on objects. We also adopted herein an event-centric schema, which can be annotated on an event-by-event basis and applied as described in Section IV-B.

**TABLE 5.** Comparison with existing studies.

| | Ability | VirtualHome [32] | HomeOntology [35] | VirtualHome2KG |
|---|---|---|---|---|
| Knowledge representation | Activity | ✓ | ✓ | ✓ |
| | Action | ✓ | ✓ | ✓ |
| | Event | | | ✓ |
| | Object | ✓ | ✓ | ✓ |
| | State of object | | ✓ | ✓ |
| | Property of object | | ✓ | ✓ |
| | Spatial data and spatial relationship | | ✓ | ✓ |
| | Temporal data and temporal relationship | | | ✓ |
| | Distinguishing properties, attributes, and affordances of objects | | | ✓ |
| Querying | Objects appearing in activities | ✓ | ✓ | ✓ |
| | OWL reasoning | | ✓ | ✓ |
| | Context of actions | | ✓ | ✓ |
| | Context of object states and spatial relations | | | ✓ |
| | Temporal duration of activity and action | | | ✓ |

The use cases examined in this study allowed us identify several problems that must be addressed to facilitate the VirtualHome2KG implementation. We classified these problems as (i) simulation platform problems, (ii) KG schema problems, and (iii) lack of human activity information (Figure 12). The subsequent subsections will discuss the causes of these problems, corresponding solutions, and potential extensions of this study.

### A. SIMULATION PLATFORM

Simulation platforms like VirtualHome, which have a one-on-one relationship between the action names and the motion data, can deterministically generate videos and simulation results based on the input order. The advantage of this platform is that a large amount of annotated data can be prepared without using activity recognition methods. However, the cost of preparing the motion data corresponding to actions can be a disadvantage. For example, in the fall risk domain discussed herein, several necessary actions were missing in the VirtualHome dataset described in Section IV-B, (e.g., "stand on one leg," "walk backwards," and "lean"). Thus, more motion data are desirable for the daily living environment, primarily in the safety domain. We expect that novel 2D video to 3D motion technologies will significantly reduce this cost.

Support is also required for multiple simultaneous actions in the daily living domain because various activities (e.g., walking while operating a smartphone) are frequently performed in real-world environments. A possible workaround for this problem is to implement separate upper and lower body motions.

Support for real-time physics simulations is also desirable because simulating uncertain events without such simulations (e.g., objects falling or people tripping over objects) is difficult.

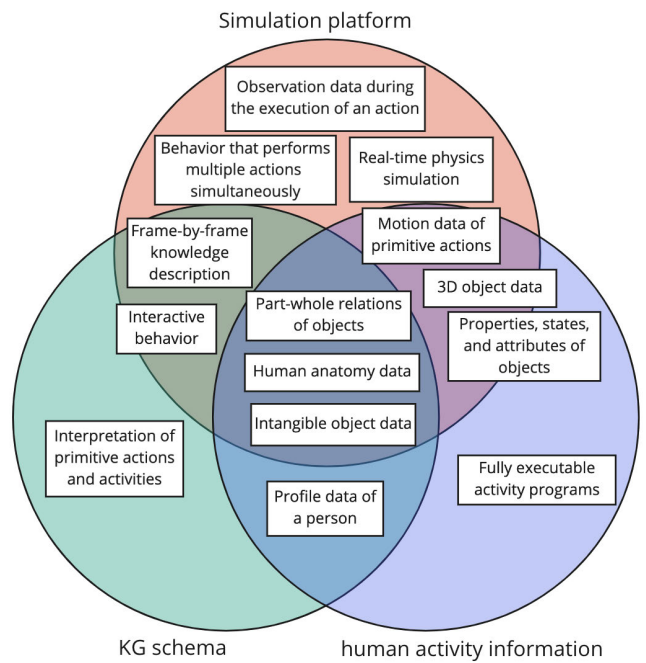

**FIGURE 12.** Problems to be addressed to apply synthetic knowledge graphs (KGs) to support daily life activities.

### B. KG SCHEMA

In addition to increasing the amount of motion data for actions, new vocabularies for actions in the KG are also required. However, further investigation is needed to determine which actions should be added because actions may have different interpretations. For example, although "stand on one leg" is intuitively considered as a primitive action, the difference between this and "lift one leg" is unclear. In other words, the interpretation depends on the observer. According to the Primitive Action Ontology [41], the definition of a primitive action excludes "actions that include interpretations of the observer" and "actions that are specialized by their target objects." According to these definitions, "stand on one leg" should be represented as a primitive action of "lift," and one (left or right) leg should be represented as a target object separated from the action. "Stand on one leg" can be also interpreted as a state caused by other actions. For example, a person always stands on one leg while walking or climbing. Some actions can be interpreted differently; hence, the primitive action definitions must be specialized for the proposed framework.

Our study focused on people who live alone. However, the schema must be extended to represent both human-human and human-robot interactions. In the current schema, people and home objects can become the target objects of events. In addition, actions involving the interactions between multiple agents can be handled by creating the `Event`, `Situation`, and `State` nodes for each agent to avoid the time-series inconsistencies in environmental information. However, a trade-off exists because the representation becomes redundant, and the data size significantly increases. Thus, further research is required to efficiently extend the schema to represent the actions and interactions of multiple agents.

### C. HUMAN ACTIVITY INFORMATION

As described in Section III-D, many of the crowdsourced "programs" in VirtualHome could not be executed using the Unity simulator. A common reason for this is the missing steps required to execute an activity. For example, if an activity contained the "SWITCHON tv (106)" step, the agent must be close to the television at that time to execute the action. Thus, "WALK tv (106)" had to appear as a prior step, and any activity lacking this step could not be executed. Another common case is an activity that specified objects with no corresponding 3D assets in the Unity model. This type could not be executed. Although a method for predicting the missing steps in an activity using a graph neural network was previously proposed [48], it has not yet demonstrated sufficient accuracy for practical application. Quality control is also required for crowdsourcing [49] executable "programs."

Information regarding human-environment interactions is required to realize more advanced applications. For example, some of the fall risks identified by a domain specialist could not be represented due to the lack of part-whole relationships between the objects in a synthetic KG (e.g., chairs and armrests). One strategy of handling this information is to interpret a chair and an armrest as a single object (e.g., an "armchair") and define "put" as the armchair object affordance. We believe that the following approaches can complement for the missing relations: (i) altering the representation to represent the missing relation; (ii) extracting the missing relation from external datasets; and (iii) collecting new data through crowdsourcing. Another problem is the information on intangible objects (e.g., a "gap" or a "hole"). Whether or not a gap or a hole is an actual entity should be treated with philosophical consideration [50], [51], which is beyond the scope of this study. From an engineering perspective, if a gap or a hole is defined as an actual entity, obtaining data pertaining to it from VirtualHome will be difficult. Alternatively, if "There is a gap between a bed and the wall" is reworded as "The bed is away from (not close to) the wall," then we can explicitly describe an inference rule for detecting the fall risk between the wall and the bed.

These points were the lessons learned through the use cases presented in this study. Additional challenges may be observed through further applications in other domains. However, we believe that our proposed VirtualHome2KG framework can sufficiently be applied to other indoor activity domains, including shopping, restaurants, and manufacturing. We also believe that the solution strategies described in this section can be used when applying synthetic KGs in other domains.

### D. FUSION OF THE KGs OF THE PHYSICAL AND VIRTUAL SPACES

We believe that the data from both physical and a virtual spaces must be combined in the future to analyze daily living activities. The three following approaches of combining both datum types are possible:

(1) considering applications for the physical space based on the KG of the simulation results on the virtual space;
(2) pre-training of machine learning models using the KG of the simulation results in the virtual space and applying them to the prediction and recognition tasks in the physical space; and
(3) graph matching between the KGs representing the contents of real and virtual videos.

Fig. 1 depicts the first approach. For example, a 3D model reproducing a physical space environment was created through 3D computer graphics software. Daily life can be simulated, and KGs can be generated by introducing the reproduced 3D model into VirtualHome2KG. This approach allows various analyses, including the detection of accident risks, such as those represented in this paper. The second approach pre-trains machine learning models for activity recognition based on the annotation data generated by VirtualHome2KG. Fine-tuning based on the dataset is expected to improve the recognition accuracy on real data in the physical space. The third approach generates KGs in both the physical and virtual spaces. KGs representing the contents of daily life can be generated from real video data using scene graph generation [3], [52] and KG construction technologies. Thus, in the future, dangerous situations are expected to be detected by graph matching between real and synthesized KGs.

## VI. CONCLUSION

In this work, we proposed the VirtualHome2KG framework to generate synthetic KGs based on an event-centric schema by simulating the daily activities in the home environment using a 3D virtual space. The proposed framework can provide source datasets for the pre-training of machine learning tasks containing more semantic information than typical annotation data because it can simultaneously generate video data and a KG representing the simulation results. It also has the potential to improve the accuracies of activity recognition, VQA, and EQA tasks because it can flexibly change the environmental conditions and augment the training data in various contexts.

We evaluated the proposed VirtualHome2KG framework in several use cases by analyzing the daily activities by querying, embedding, and clustering and detecting the fall risks among older adults based on expert knowledge. These use cases provided us with promising results that can help maintain human safety in daily life and improve the quality of life.

We also identified and classified the lessons learned through this study and discussed the potential solutions for implementing advanced applications to support daily life. In the future, as cyber-physical systems and digital twin technologies become widely implemented in serviced housing for older adults, we believe that the use cases we presented here can be developed into AI system applications to warn residents about risky behaviors and offer alternatives.

We plan to extend the proposed VirtualHome2KG framework by addressing the challenges described in Section V and providing additional data resources for practical applications by reproducing physical environments in the virtual space.

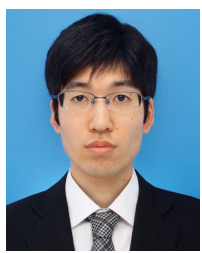

**SHUSAKU EGAMI** received the Ph.D. degree in engineering from The University of Electro-Communications, Tokyo, Japan, in 2019. He is currently a Researcher with the National Institute of Advanced Industrial Science and Technology, Japan. He is also a part-time Lecturer with Hosei University, Tokyo, and a collaborative Researcher with The University of Electro-Communications. His research interests include semantic web, ontologies, and knowledge graphs.

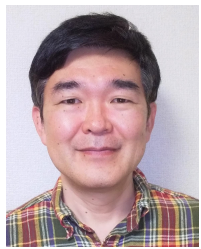

**TAKANORI UGAI** received the Ph.D. degree in engineering from Tokyo Institute of Technology, Tokyo, Japan, in 2013. Since 1992, he has been a Researcher with Fujitsu Ltd., Japan. He is also a Researcher with the National Institute of Advanced Industrial Science and Technology, Japan, and a Lecturer with Tsukuba University, Ibaraki, Japan. His research interests include knowledge graphs, graph neural networks, and requirements engineering.

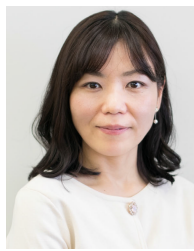

**MIKIKO OONO** received the M.P.H. degree from San Jose State University, San Jose, CA, USA, with a focus on community health education, and the Ph.D. degree from Kobe University, in 2013. She received the Certification of Public Health, in 2009. She is currently a Researcher with the Artificial Intelligence Research Center, National Institute of Advanced Industrial Science and Technology (AIST), Japan. Her research interests include injury prevention and health promotion using an engineering approach.

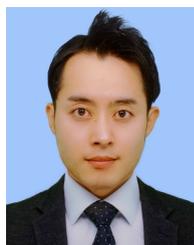

**KOJI KITAMURA** received the Ph.D. degree from the Graduate School of Mechanical Engineering, Tokyo University of Science, in 2008. He joined the Digital Human Research Center, National Institute of Advanced Industrial Science and Technology (AIST), Japan, in 2008. He is currently a member of the Living Activity Modeling Research Team, Artificial Intelligence Research Center, AIST. His research interests include safety and active everyday life based on sensing systems, data mining technologies, human behavior modeling technologies, and AI technologies. His main research interests include childhood injury prevention and elderly care support.

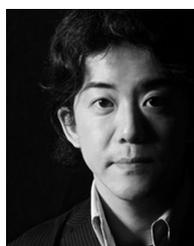

**KEN FUKUDA** received the Ph.D. degree in information science from The University of Tokyo, in 2001. He joined the National Institute of Advanced Industrial Science and Technology (AIST), as a Research Scientist, in 2001. He was a Visiting Lecturer with The University of Tokyo and a Visiting Associate Professor with Waseda University. Since then, he has led multiple national and international projects, covering a broad range of interdisciplinary research from life science to service science. He is currently leading the Data Knowledge Integration Team, Artificial Intelligence Research Center, AIST. His research interests include knowledge representation, knowledge graphs, data-knowledge integration, and human–robot interaction. As a Ph.D. student, he received the Research Fellowship for Young Scientists.